\documentclass[10pt,twocolumn,letterpaper]{article}

\usepackage{cvpr}
\usepackage{times}
\usepackage{epsfig}
\usepackage{graphicx}
\usepackage{amsmath}
\usepackage{amssymb}
\usepackage{textcomp,gensymb}

\usepackage{xcolor}
\usepackage[colorlinks = true,
            linkcolor = black,
            urlcolor  = blue,
            citecolor = black,
            anchorcolor = black]{hyperref}

\newcommand{\MYhref}[3][blue]{\href{#2}{\color{#1}{#3}}}

\cvprfinalcopy 

\def\cvprPaperID{****} 

\begin{document}

\title{Simulating Structure-from-Motion}

\author{Martin Hahner\\
Exchange Student D-INFK\\
ITU Copenhagen\\
\and
Panagiotis Bountouris\\
Master Student D-MAVT\\
ETH Zurich\\
\and
Orestis Varesis\\
Master Student D-MAVT\\
ETH Zurich\\
}

\maketitle

\begin{abstract}
The implementation of a Structure-from-Motion (SfM) pipeline from a synthetically generated scene as well as the investigation of the faithfulness of diverse reconstructions is the subject of this project. A series of different SfM reconstructions are implemented and their camera pose estimations are being contrasted with their respective ground truth locations. Finally, injection of ground truth location data into the rendered images in order to reduce the estimation error of the camera poses is studied as well.
\end{abstract}

\section{Introduction} \label{Intro}
Structure-from-Motion (SfM) is a precious tool for numerous applications
such as localization and tracking.

\subsection{Problem Statement} \label{probState}
Currently, SfM algorithms are mostly based on real world data, which however
are often expensive and limited to reproduce. Those limitations of real world data affect how state-of-the-art
algorithms behave and thus limit their range of applications. This problem is tackled in this project,
in which the data are derived from a synthetically generated scene and thus are easier
to reproduce. Synthetically generated scenes can be constructed for a wide range of applications varying from indoor scenes to outdoor scenes such as complete city structures. 

In addition, with the knowledge of the exact ground truth data we are able to tackle the problem of validation of different SfM reconstruction algorithms.

\subsection{Related Work} \label{relWork}
Most of the already implemented work regarding Structure-from-Motion reconstructions are based on images captured in the real world. Such algorithms can be found in the work from J. Ventura~\cite{Ventura2016} where the data are collected from a camera which rotates in a spherical motion. The data are collected from real world scenes and untrained users. Another SfM reconstruction related work is from J. L. Sch\"onberger and J.-M. Frahm~\cite{schoenberger2016sfm}. In this article, a new SfM technique was presented which improved the accuracy and robustness of the reconstruction following a proposed full SfM reconstruction pipeline. The data were again derived from user taken images found on the Internet.
In the work of Z. Zhang et al.~\cite{7487210} a synthetically generated urban environment as well as an indoor scene are implemented and images are rendered with different choices of camera parameters in order to determine the impact on the accuracy and robustness of the implemented algorithm.

\subsection{Project Contribution} \label{projContrib}
The  majority of the related work as shown above has focused on real world data, thus, a fairly unexplored field is the Structure-from-Motion reconstruction from synthetically generated data which comes as a natural extension to the aforementioned work. In chapter~\ref{projDesc} we discuss the main workload of the project which mainly consists of the generation of a synthetic scene, generation of a series of SfM reconstructions and the calibration of the camera poses between the ground truth and the reconstructions. In chapter~\ref{Results}, the results of the different SfM reconstructions and diagrams of the calibration metrics of the camera poses are being shown. In chapter~\ref{Conclusions} conclusions regarding our project are made and in chapter~\ref{FutureWork} suggestions for future work are proposed.

\subsection{Member Contribution} \label{membContrib}
This project consists mainly of five parts, namely: first, the investigation of relevant literature in order to find synthetically generated scenes and state-of-the-art SfM reconstruction algorithms. Second, the rendering of images from the synthetic scene. Third, running an extensive amount of SfM reconstructions using the rendered images and fourth, the proposal of an appropriate method to validate the generated SfM reconstructions with regard to the ground truth.

The first two parts of the project were equally distributed among all three group members. For the third part of the project, running many SfM reconstructions all members contributed with Martin Hahner having the main contribution. Then, the validation of the generated SfM reconstructions which consists of the extraction and calibration of the camera poses between the ground truth and the generated SfM reconstructions was mainly realized by the group members Panagiotis Bountouris and Orestis Varesis. 

Finally, in parallel with the validation part, one extra step with respect to the initial project proposal was conducted. The attempt to improve the SfM reconstructions by inserting the ground truth camera location data to the EXIF data of the rendered images before running SfM reconstructions with them. For this additional task again, all members contributed with Martin Hahner having the main contribution.

\section{Project Description} \label{projDesc}
In this chapter the project pipeline as well as some remarks regarding the goals of the project are presented.

\subsection{Synthetically Generated Scene} \label{synthScene}
The synthetically generated scene which is used in this project is a synthetic dataset representing a typical urban environment of a city that was provided by the University of Zurich~\cite{7487210}. The outdoor scene was implemented in blender and in figure 1 two snapshots of the scene from different viewpoints are shown. A closer look to figure 1 reveals the black trajectory which the camera follows on both of the snapshots.

\begin{figure}[ht] \label{figsynthScene}
	\begin{center}
      \includegraphics[width=0.4\linewidth]{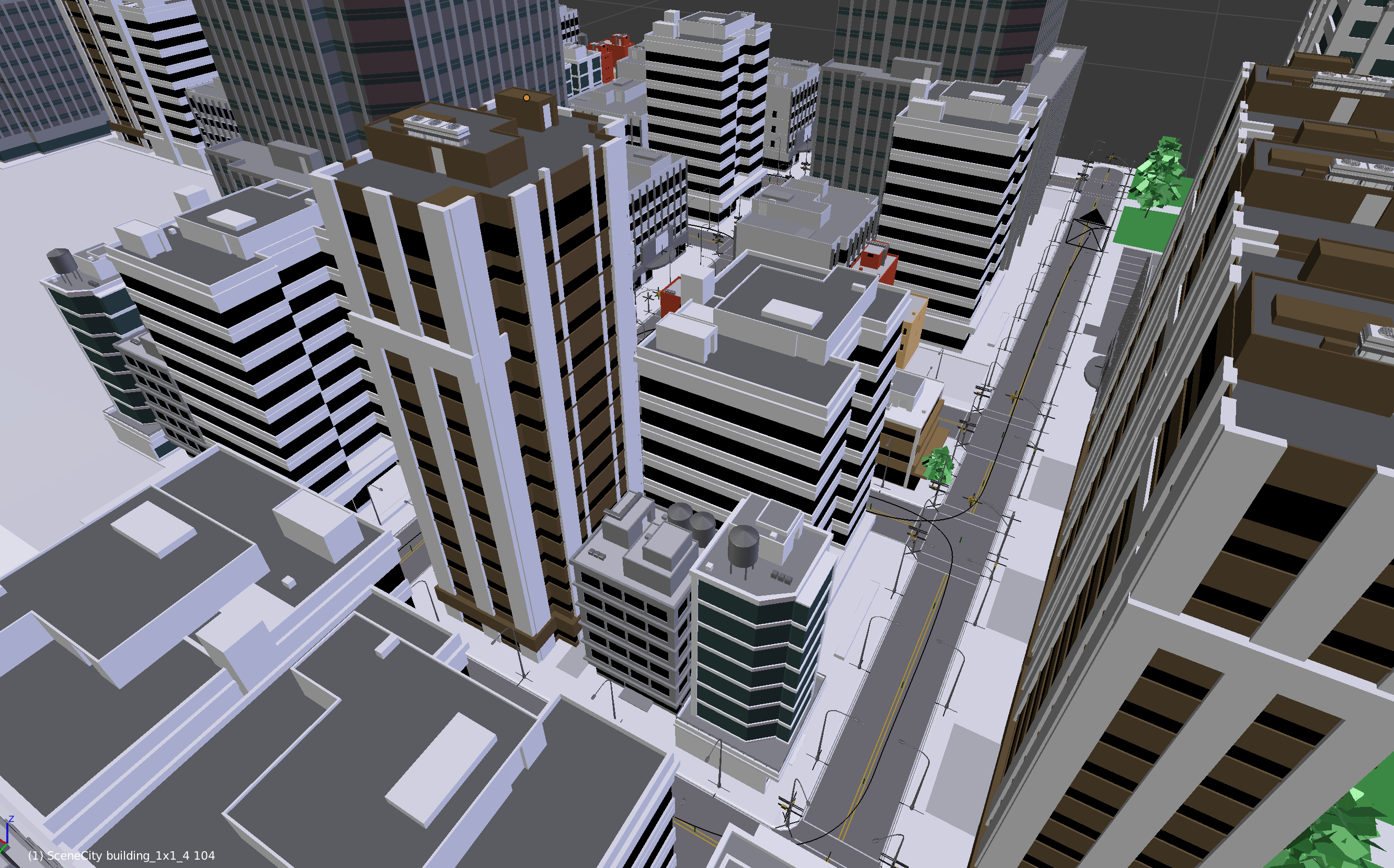}
      \includegraphics[width=0.4\linewidth]{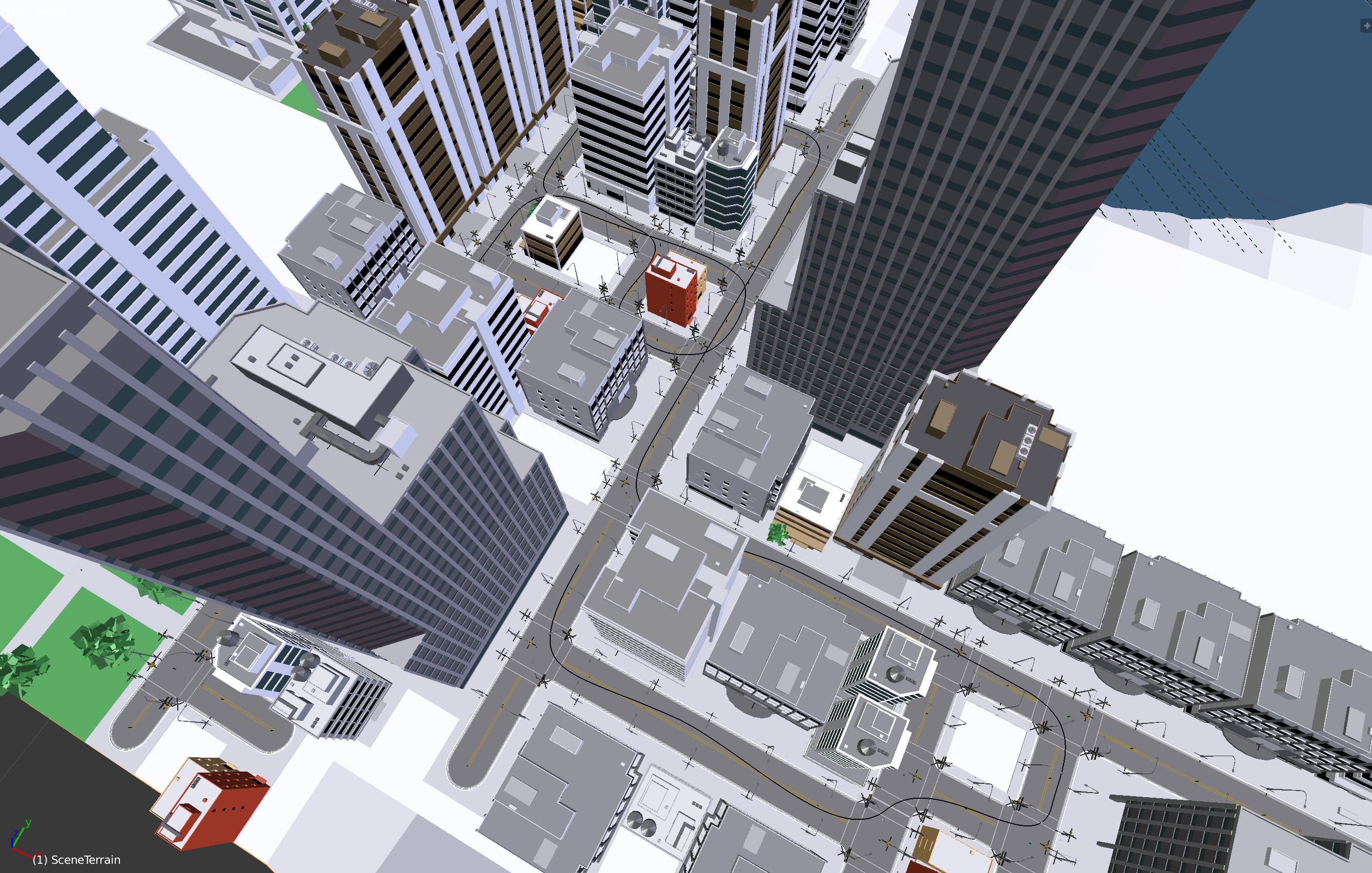}
	\end{center}
   \caption{synthetically generated outdoor scene}
\end{figure}

The camera trajectory consists of 3000 frames in total and the camera specifications can be altered in order to have the best reconstruction possible. After several attempts, for best results the images were rendered for each frame and for completeness of this project three different camera orientations (forward, non-tilted sideview and tilted sideview) were investigated. Thereof, only the forward motion renderings were provided by the authors of paper~\cite{7487210}, for sideview motion the camera field of view was increased from $\sim$90$\degree$ to 140$\degree$ in order to get more distinct features per image as shown in figure 2.

\begin{figure}[ht] \label{figrenderings}
    \begin{center}
       \includegraphics[width=0.3\linewidth]{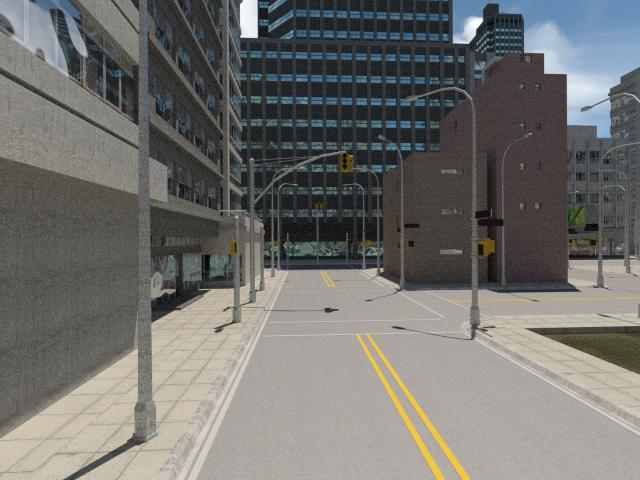}
       \includegraphics[width=0.3\linewidth]{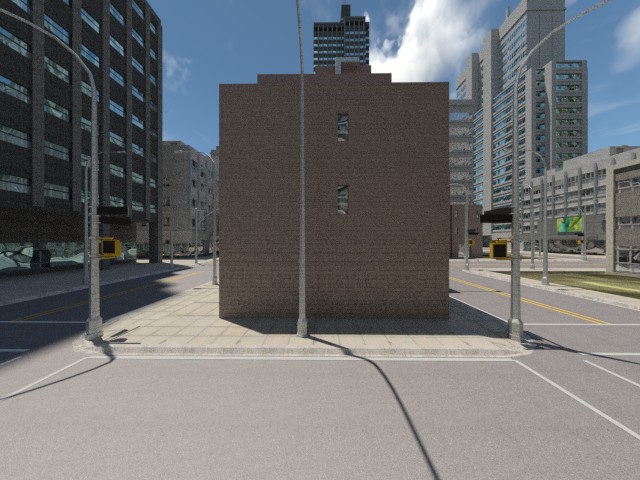}
       \includegraphics[width=0.3\linewidth]{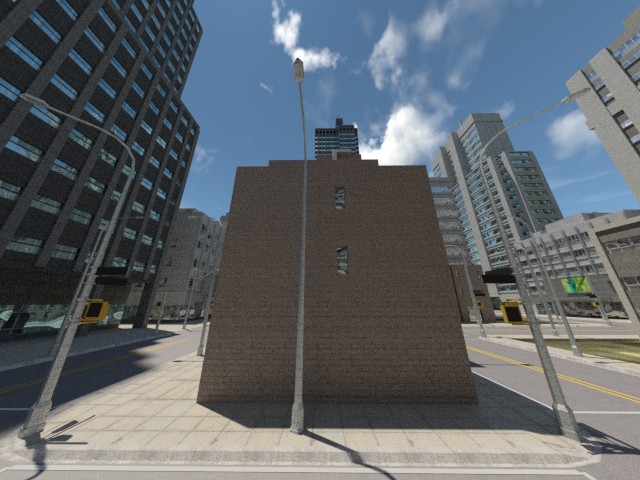}
    \end{center}
	\caption{image renderings for (a): forward motion, (b): non-tilted sideview motion, (c): tilted sideview motion}
\end{figure}

Looking at the computation time of those image renderings, each image took on average around two minutes to render on a HP Z 440 workstation with 32GB RAM, a 3.5 GHz Intel Xeon processor and one NVIDIA QUADRO K4200 graphics card. More than 200 hours were needed, only to render all the images for the two modified (sideview) datasets.

\subsection{SfM Reconstruction with Sequential Feature Matching} \label{SFMseq}
To compute the SfM reconstructions COLMAP~\cite{schoenberger2016mvs} was used. COLMAP is open-source, uses a graphical as well as a command-line interface and serves as a pipeline for Structure-from-Motion and Multi-View-Stereo (MVS). It offers a lot of control through a large variety of parameters. 

The rendered images described above were fed into COLMAP. COLMAP then first looks for distinguishable features in each image separately as seen in figure 3 (a) and after that they have to be matched with features from other images as seen in figure 4. 

\begin{figure}[ht] \label{featuresFound}
	\begin{center}
      \includegraphics[width=0.45\linewidth]{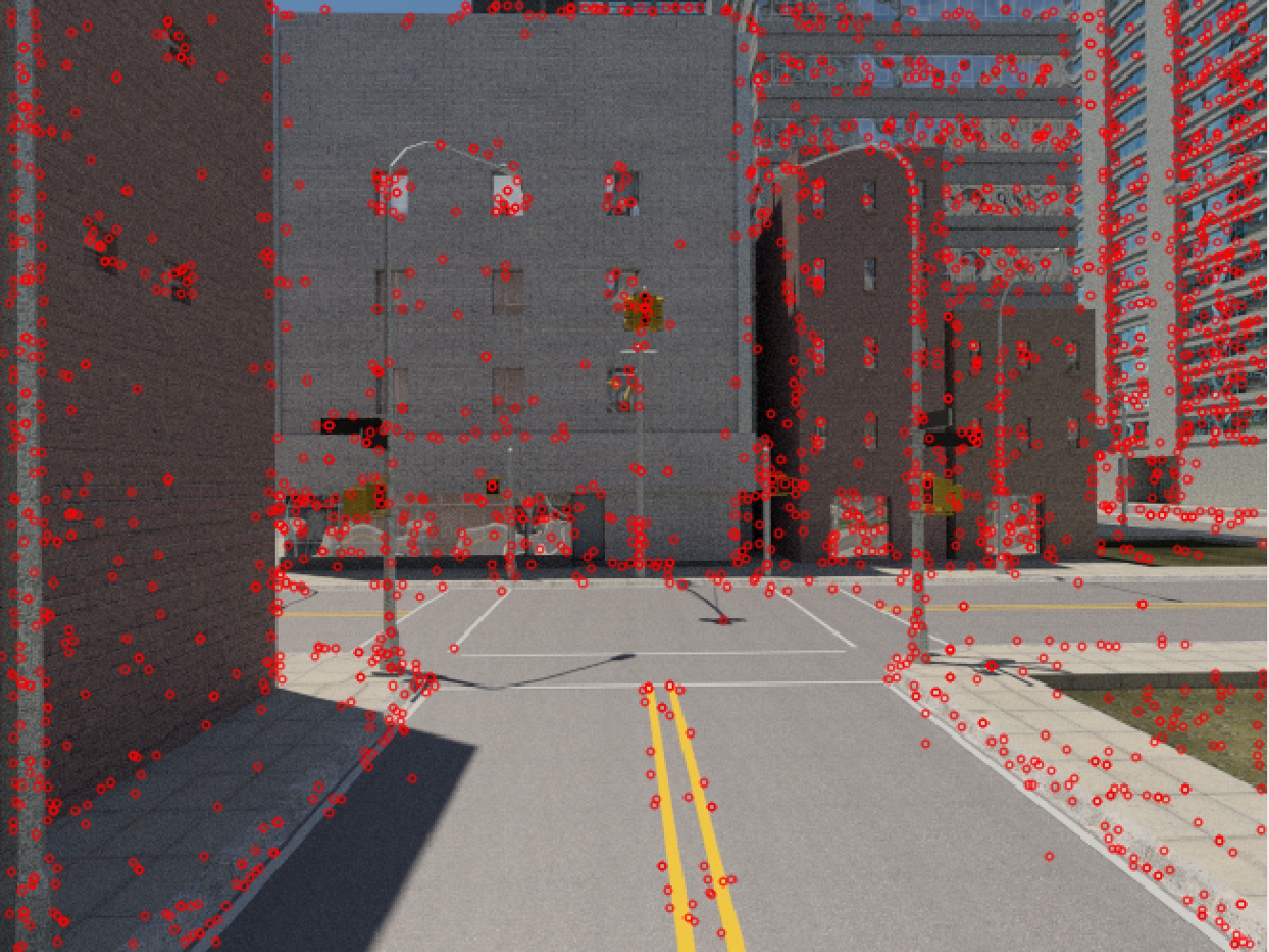}
       \includegraphics[width=0.45\linewidth]{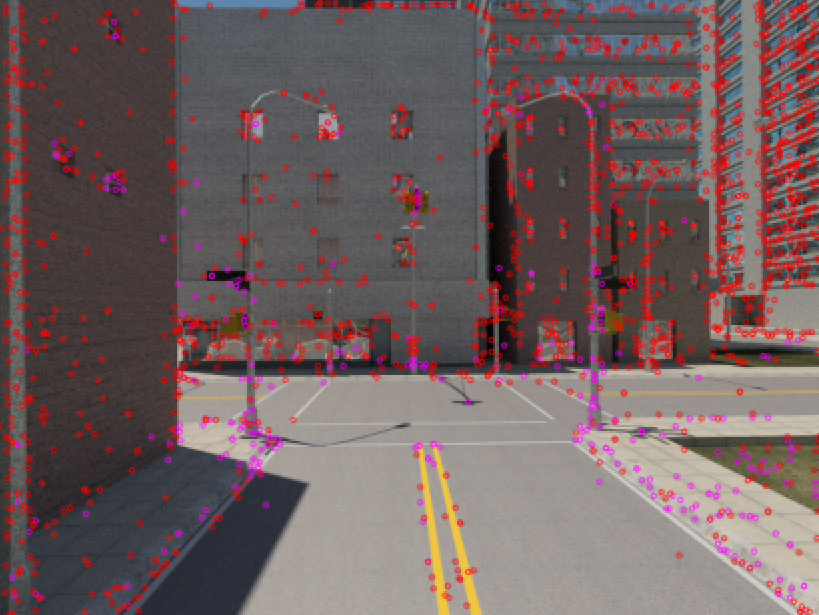}   
	\end{center}
   \caption{example image with (a): all features found, (b): all features found (in red) and all features matched (in pink)}
\end{figure}

\begin{figure}[ht] \label{featureMatchesOfTwoImages}
	\begin{center}
      \includegraphics[width=0.9\linewidth]{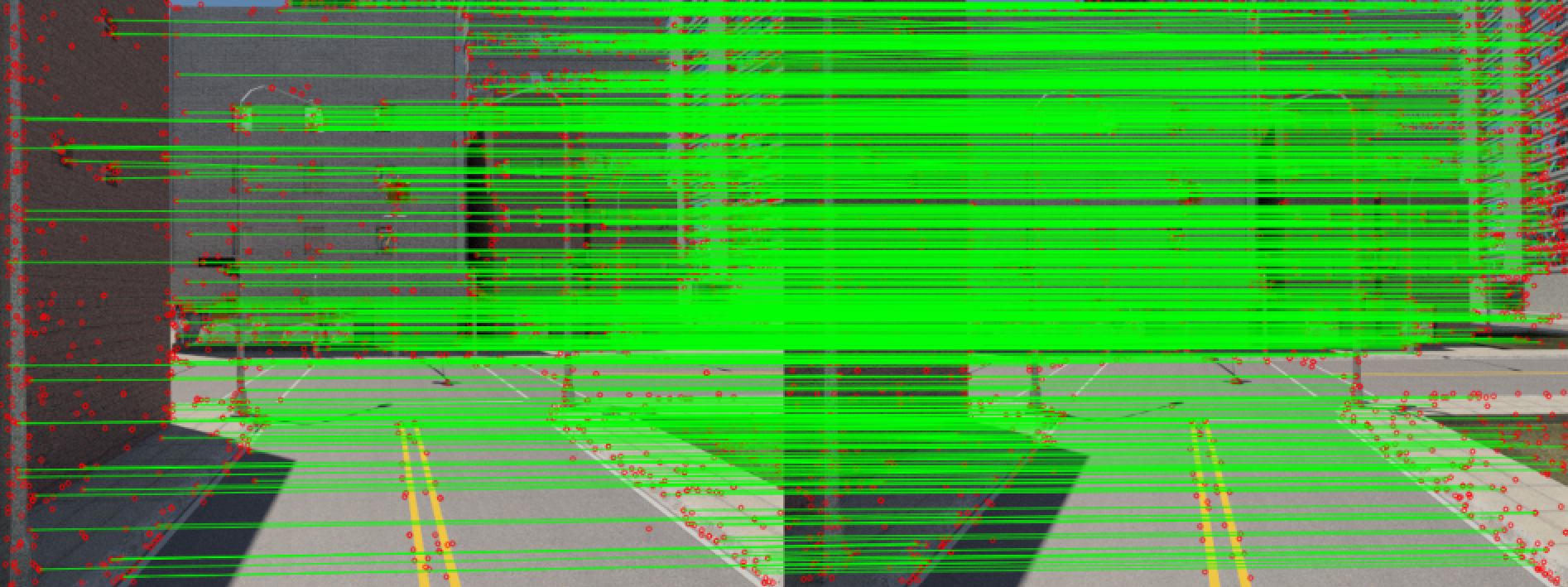}
	\end{center}
   \caption{two sequential images with all their matched features}
\end{figure}

A variety of parameters as well as the different feature matching modes COLMAP offers can be seen in figure 5. For sequential images a few distinct parameters are defined, the most important one being the overlap parameter, which specifies how many images in the image sequence are considered for feature matching. 

\begin{figure}[ht] \label{featureparams}
	\begin{center}
      \includegraphics[width=0.39\linewidth]{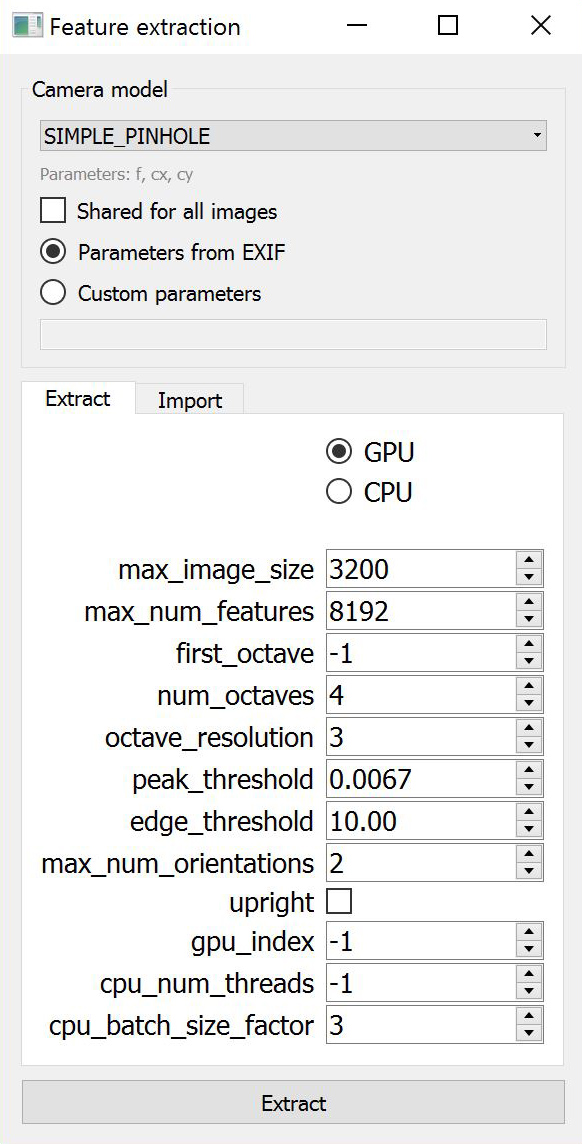}
       \includegraphics[width=0.45\linewidth]{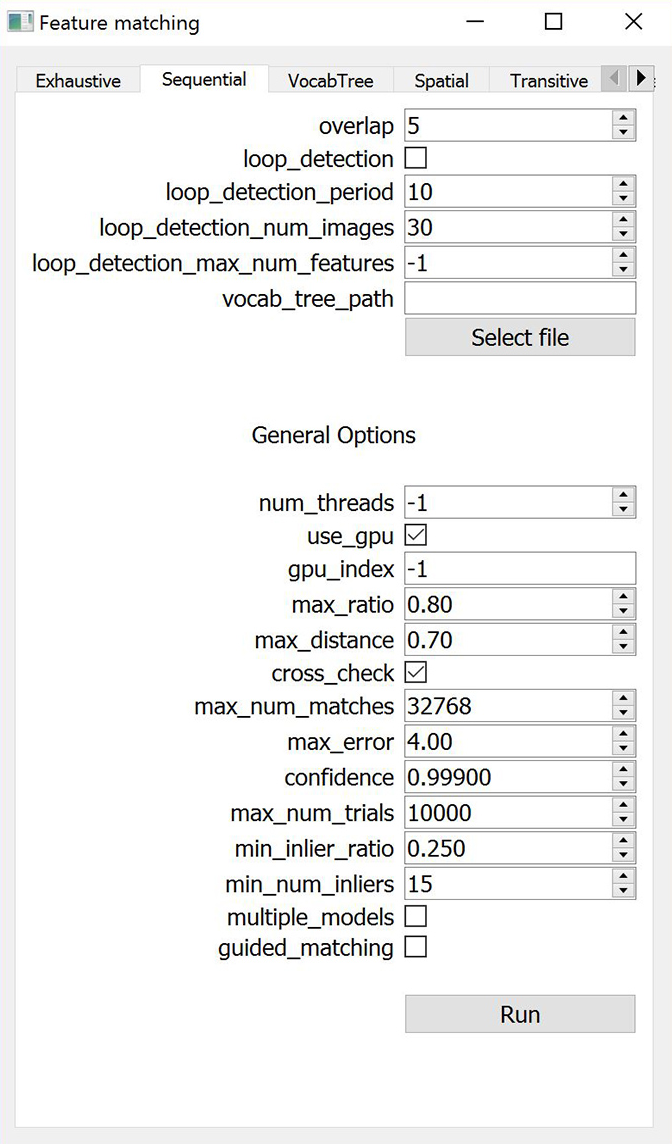}   
	\end{center}
   \caption{default COLMAP parameters for (a): feature extraction, (b): feature matching}
\end{figure}

For this project the overlap parameter was chosen to be ten sequential images, since this value gave a vast amount of feature matches (mostly between 400 and 1000 feature matches per image pair).

After having the feature matches between all the 3000 sequential frames taken from the trajectory in blender, an SfM reconstruction is computed. Three different reconstructions are shown in figure 6.

\begin{figure}[ht] \label{SfMreconstruction}
	\begin{center}
       \includegraphics[width=0.45\linewidth]{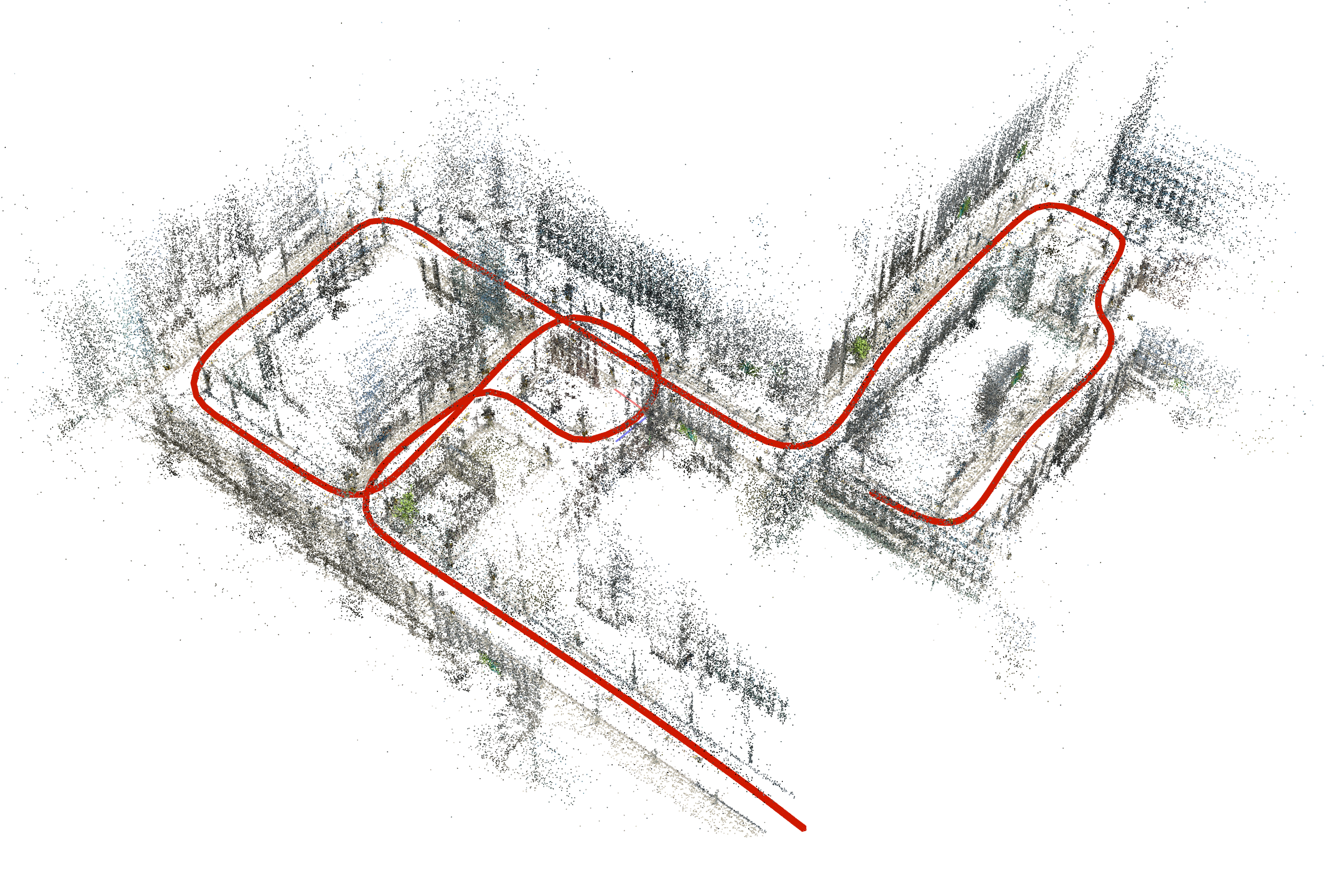}
       \includegraphics[width=0.45\linewidth]{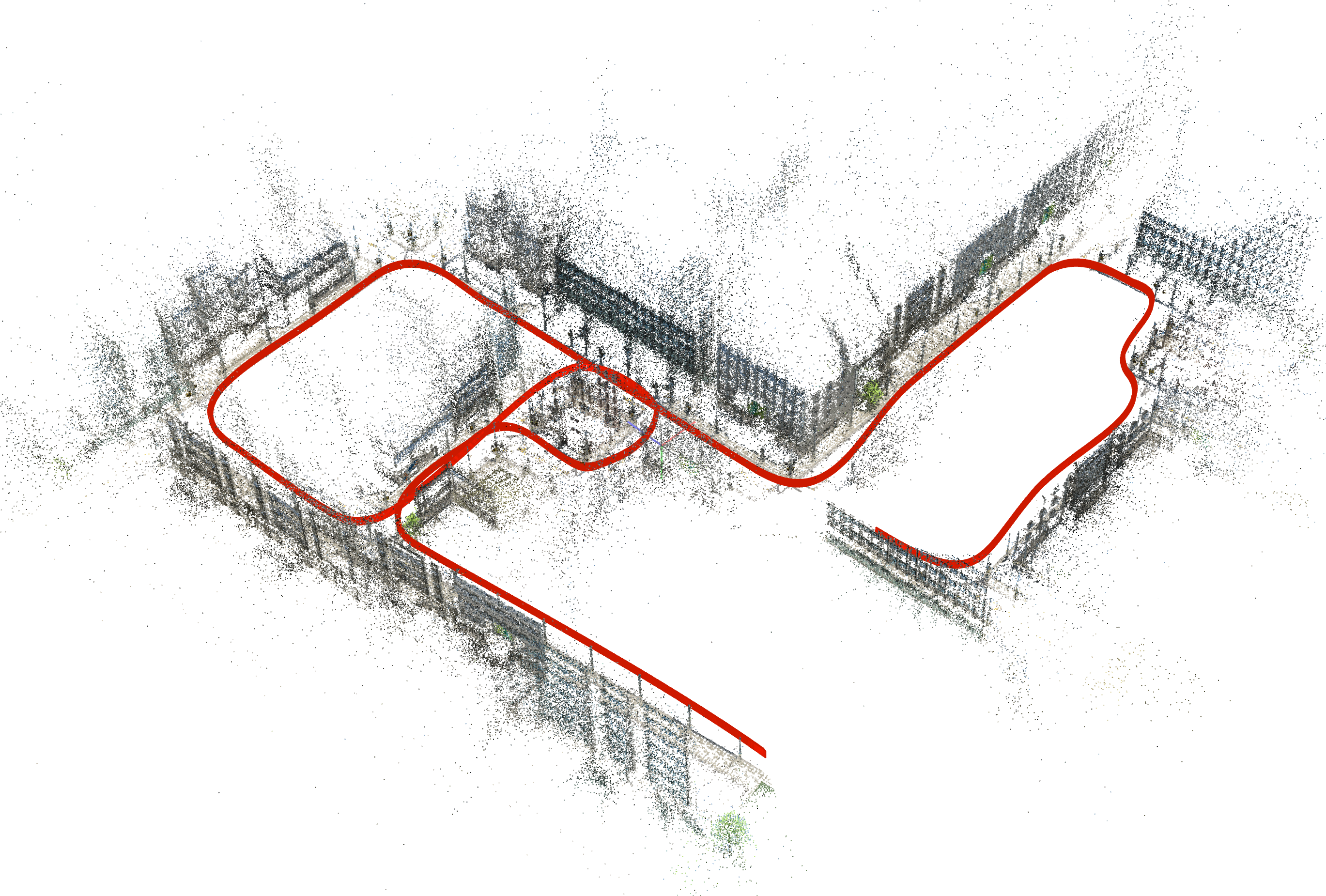}
       \includegraphics[width=0.45\linewidth]{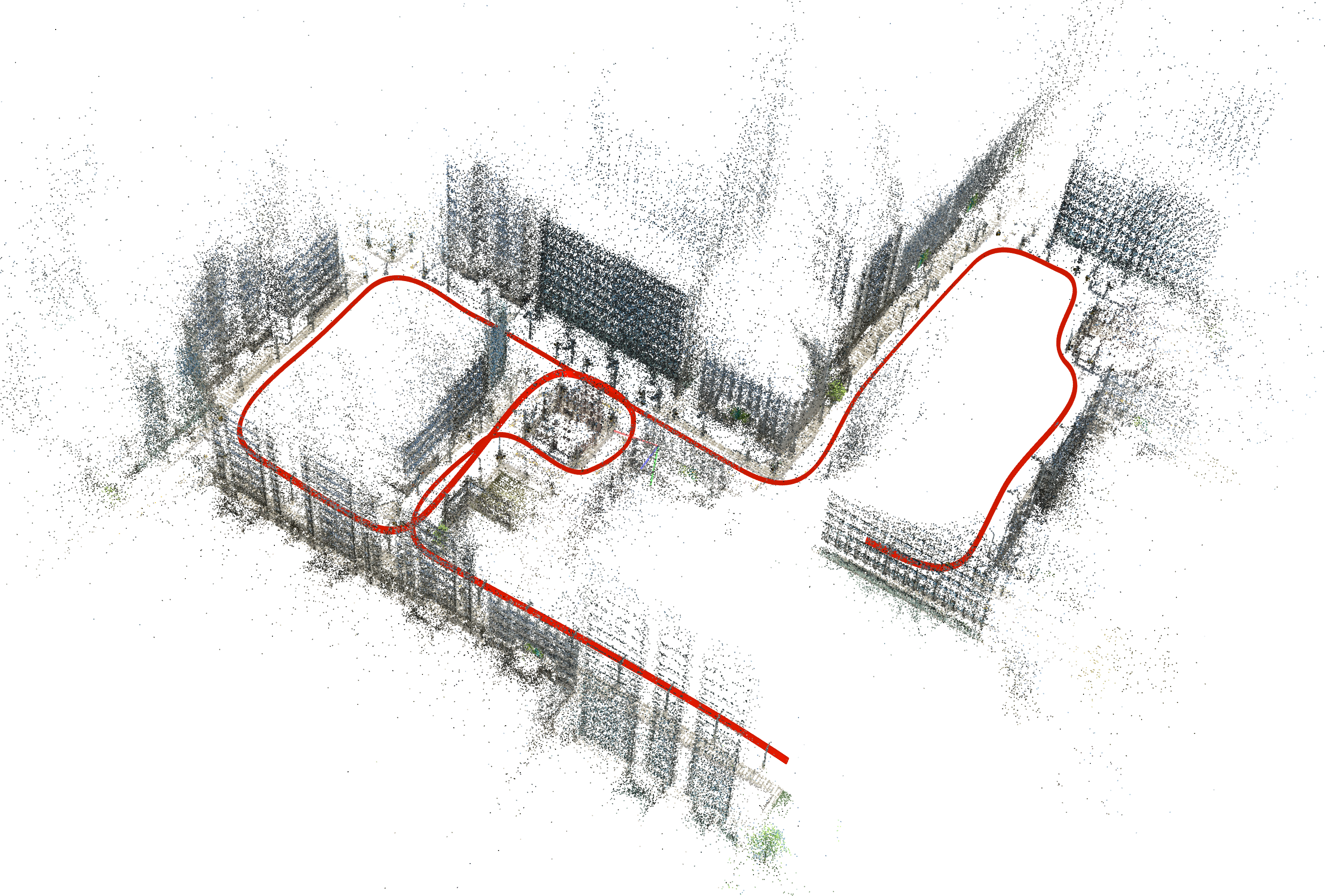}
       \caption{SfM reconstruction for (a): forward motion, (b): non-tilted sideview motion, (c): tilted sideview motion}
	\end{center}
\end{figure}

Comparing the three different reconstructions, the tilted version, even though it has less points than the non-tilted reconstruction, which means it has less feature matches between its images, achieved the overall best representation of the 3D synthetic scene. This is explained due to the tilted camera orientation. The camera can see more of the building's facade and therefore can represent the points in the SfM reconstruction in a more balanced manner. This also results in a much smaller drift in contrast to the other two reconstructions. The forward facing reconstruction had the worst result of the three, that is why for the rest of the project we focused more on the two sideview reconstructions.

The aforementioned process is both quite computationally expensive and time consuming. COLMAP needed roughly 11 hours to compute each SfM reconstruction with sequential feature matching using a HP Z 440 workstation with 32GB RAM, a 3.5 GHz Intel Xeon processor and one NVIDIA QUADRO K4200 graphics card.

\subsection{Calibration of Camera Poses} \label{Calib}
Once the SfM reconstructions in COLMAP were completed, the next part of the project was the calibration of the camera poses, meaning to align the ground truth camera positions from blender with the estimated camera positions from COLMAP in a common coordinate system. In figure 7, one camera pose of the trajectory is shown both in blender and in COLMAP. 

\begin{figure}[ht] \label{cameraposes}
	\begin{center}
   		\includegraphics[width=0.425\linewidth]{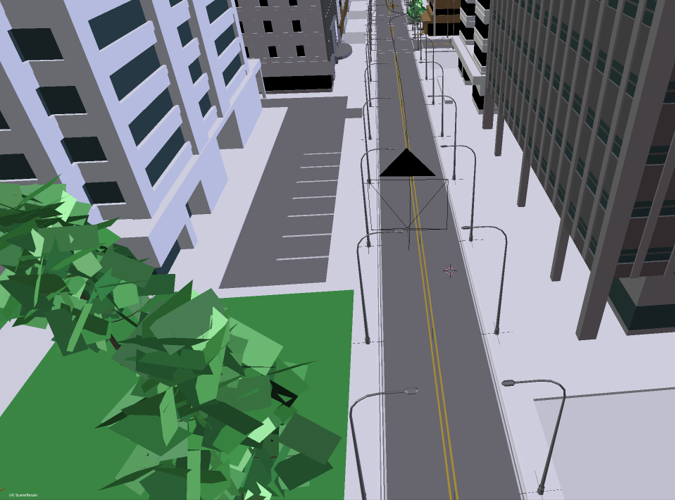}
   		\includegraphics[width=0.45\linewidth]{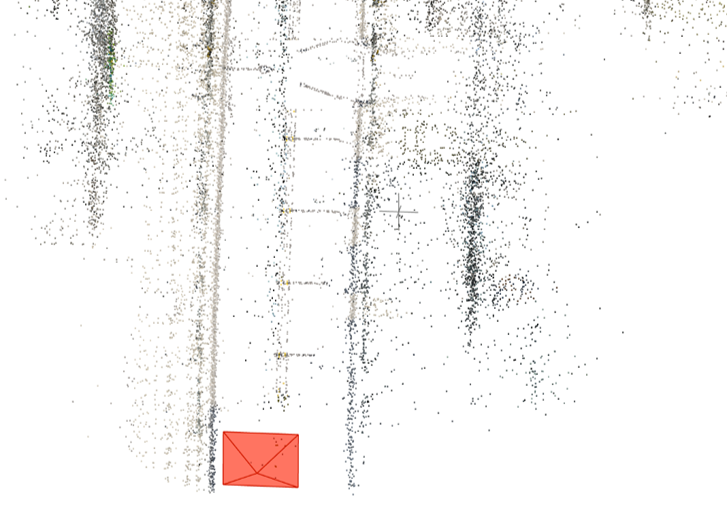}
   		\caption{camera poses in (a): blender, (b): COLMAP}
	\end{center}
\end{figure}

At this point it is worth mentioning again that the ground truth in this project was the information that was extracted from blender, so it is 100\% precise and contains no error and no noise of any sort. The way this information was extracted, is via a blender text file export that contains all the camera poses with explicit details regarding the position and orientation of each rendered image. Similarly, after the completion of the SfM reconstructions in COLMAP, a text file named "images.txt" is returned which includes all the estimated camera poses of the reconstructed images along with matched feature points. 

From the previous paragraph, it is now clear that this part of the project consisted of two subparts. First, the extraction of the camera poses from these text files and then the transformation of the two 3D point sets to a common coordinate system. Regarding the extraction of the camera poses, although the text file from blender was easily accessible and manageable in Matlab, the respective text file from COLMAP was infeasible to access and process in a similar way due to the vast amount of matched feature points the file contains as well - text file sizes of 150-400MB had to be handled. So for this file type, the information regarding the camera poses was initially extracted using Microsoft Excel by applying filters to the columns and then the Excel file instead of the original text file was accessed and processed with Matlab. So also due to the different representation of the orientation in each type of text file, two different Matlab scripts were written. In detail, the orientation of the camera poses in blender's file was represented through Euler angles, while in COLMAP's file orientation was represented with quaternions.
 
Finally, in order to get a meaningful comparison between the camera poses of the two programs, a calibration between them was needed since they were derived with respect to different coordinate systems. For this purpose, a method based on paper ~\cite{horn1987closed} and implemented by Dr. Christian Wengert and Dr. Gerald Bianchi of the Computer Vision Laboratory at ETH Zurich was used. In this implementation, given two 3D point sets a translation and rotation is computed in order to find the relative transformation between those two 3D point sets. 

As it will be stated in more detail in chapter~\ref{Results}, the camera poses of the SfM reconstructions in COLMAP presented a root-mean-square (RMS) error of approximately $1m$ compared to the ground truth in the planar $x$ and $y$ directions. Considering the overall dimension of the synthetic city scene of approximately $100m$ x $100m$, such an error could not be characterized as considerable, although a better accuracy was expected.

\subsection{SfM reconstruction with Spatial Feature Matching} \label{SFMspat}
In an attempt to improve the camera poses' errors, an extra step with respect to the initial project proposal was made. 
The key point was, to enable COLMAP to use the ground truth location data of the camera positions from blender as a starting value for its reconstruction computation. So first, the code of COLMAP was investigated and it was found that it can read GPS information from EXIF data. Then, a script was written, that added the ground truth location data from blender into the corresponding EXIF fields of each rendered image before feeding them again into COLMAP. In order to make this properly work one line of code of COLMAP also had to be changed to handle minus signs while reading the corresponding EXIF fields. Finally, the feature matching parameters that COLMAP offers  were investigated in the underlying source code and were altered in a way that led to way less, but way better feature matches as depicted in figure 8.    

\begin{figure}[ht] \label{featureMatchingParameters}
	\begin{center}
   		\includegraphics[width=0.4\linewidth]{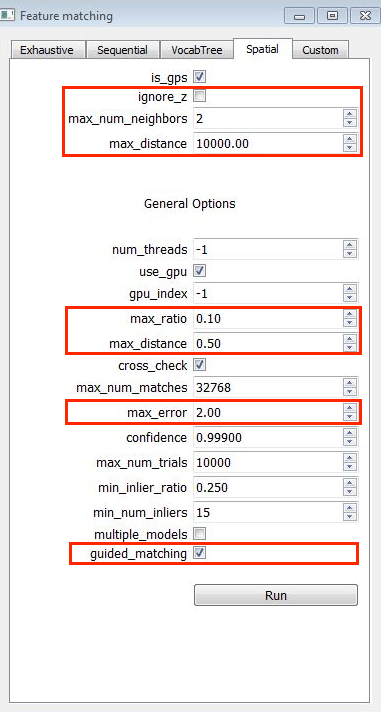}
   		\caption{presumable better feature matching parameters}
	\end{center}
\end{figure}

With those settings COLMAP needed roughly 9 hours to compute each SfM reconstruction with spatial feature matching using a HP Z 440 workstation with 32GB RAM, a 3.5 GHz Intel Xeon processor and one NVIDIA QUADRO K4200 graphics card.

\section{Results} \label{Results}
In this chapter the results of the calibration between the two camera poses of blender and COLMAP, which was introduced in the section~\ref{Calib}, are presented. The three following figures show the poses from blender (in red) and COLMAP (in cyan) for six different SfM reconstructions. All of those reconstructions followed the pipeline introduced in  chapter~\ref{projDesc}. 

Firstly, in figure 9 a comparison of the camera poses from blender and COLMAP using sequential feature matching with ten sequential images are presented.

\begin{figure}[ht] \label{cameraposesab}
	\begin{center}
   		\includegraphics[width=0.49\linewidth]{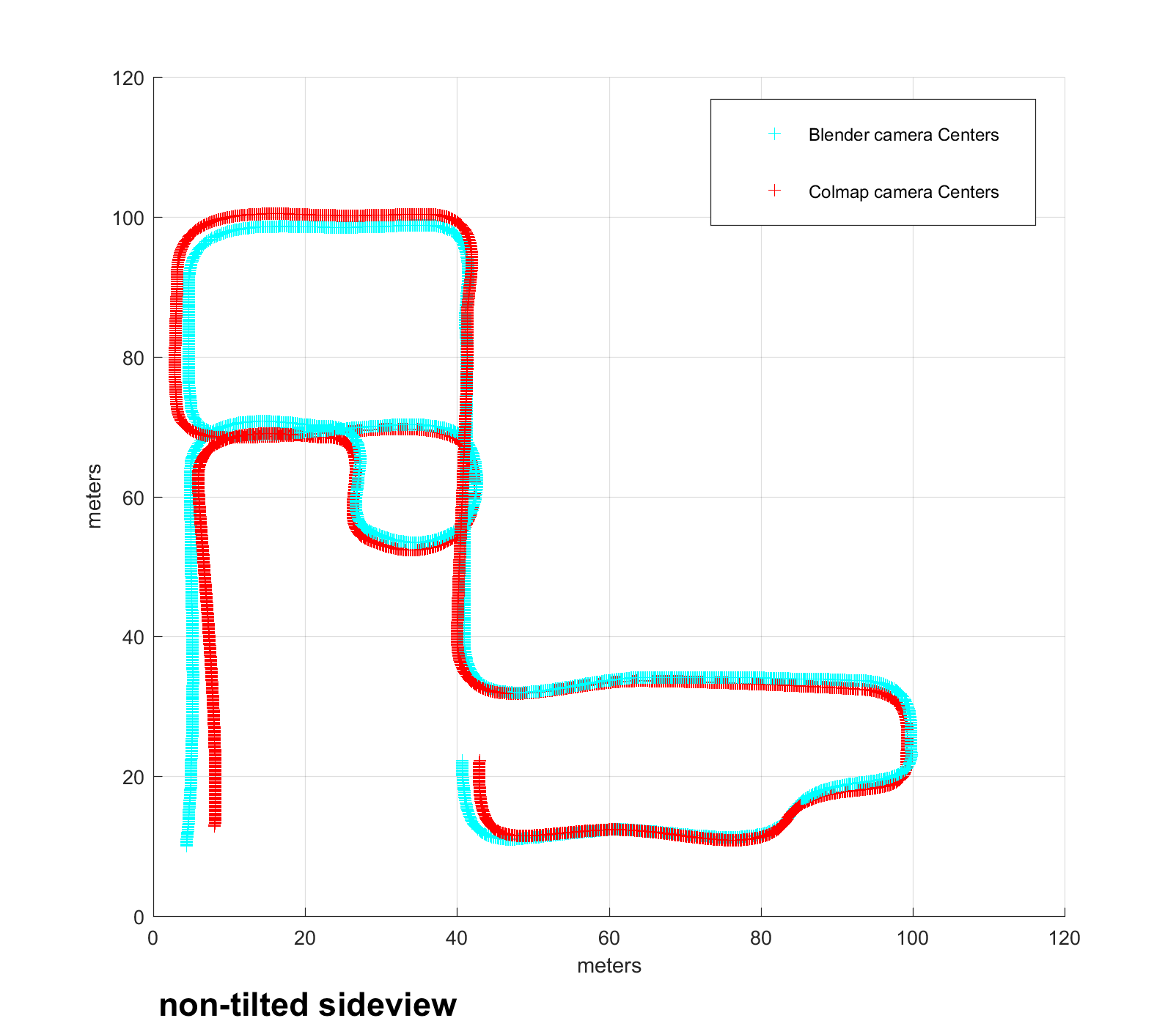}
   		\includegraphics[width=0.49\linewidth]{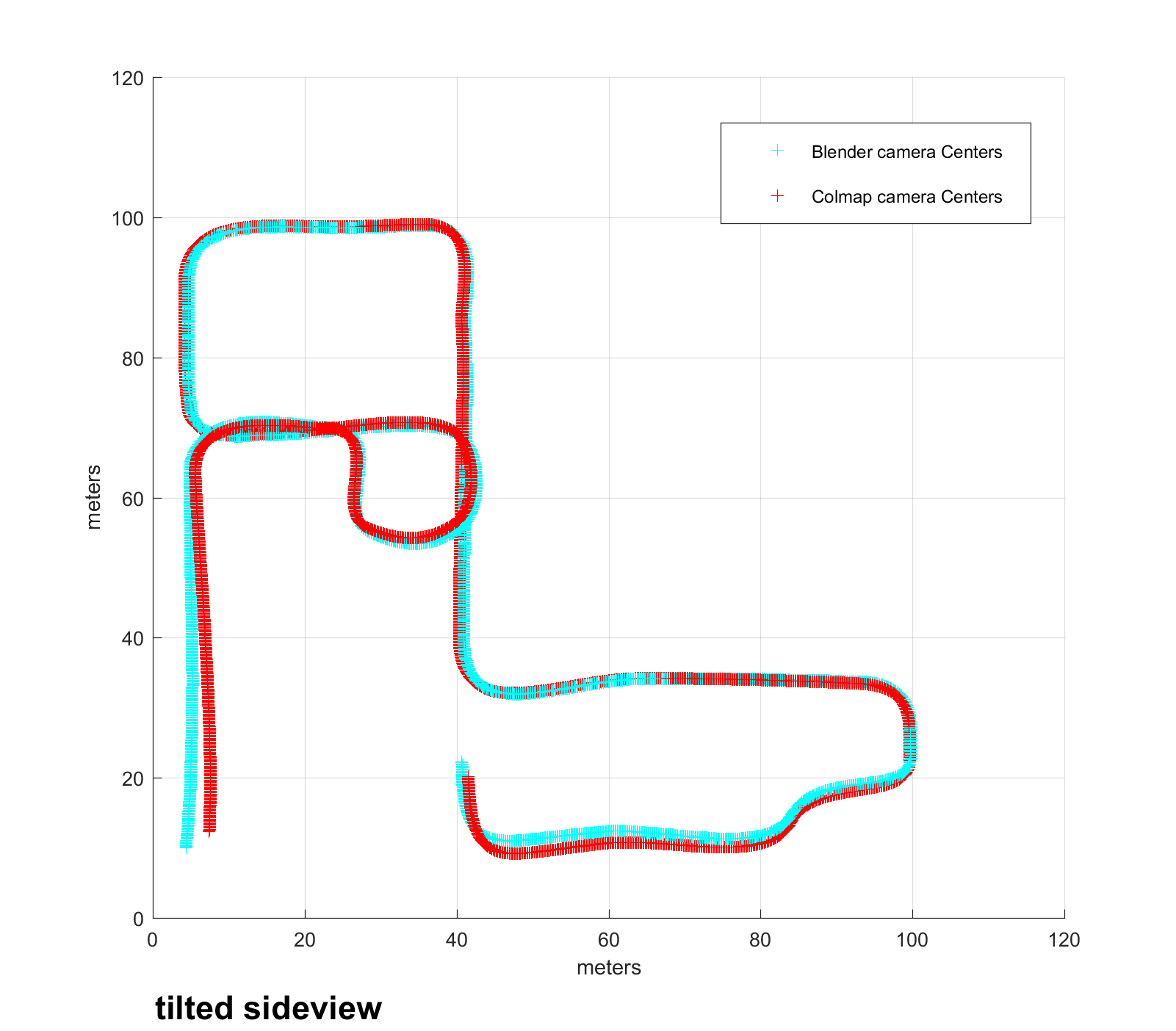}
   		\caption{camera poses of (a): non-tilted sideview, (b): tilted sideview}
	\end{center}
\end{figure}

After inserting the ground truth GPS location data into the EXIF information of the rendered images and re-computing the SfM reconstructions with spatial feature matching and ten neighboring images, a comparison of the camera poses from blender and COLMAP is presented in figure 10.

\begin{figure}[ht] \label{cameraposescd}
	\begin{center}
       	\includegraphics[width=0.49\linewidth]{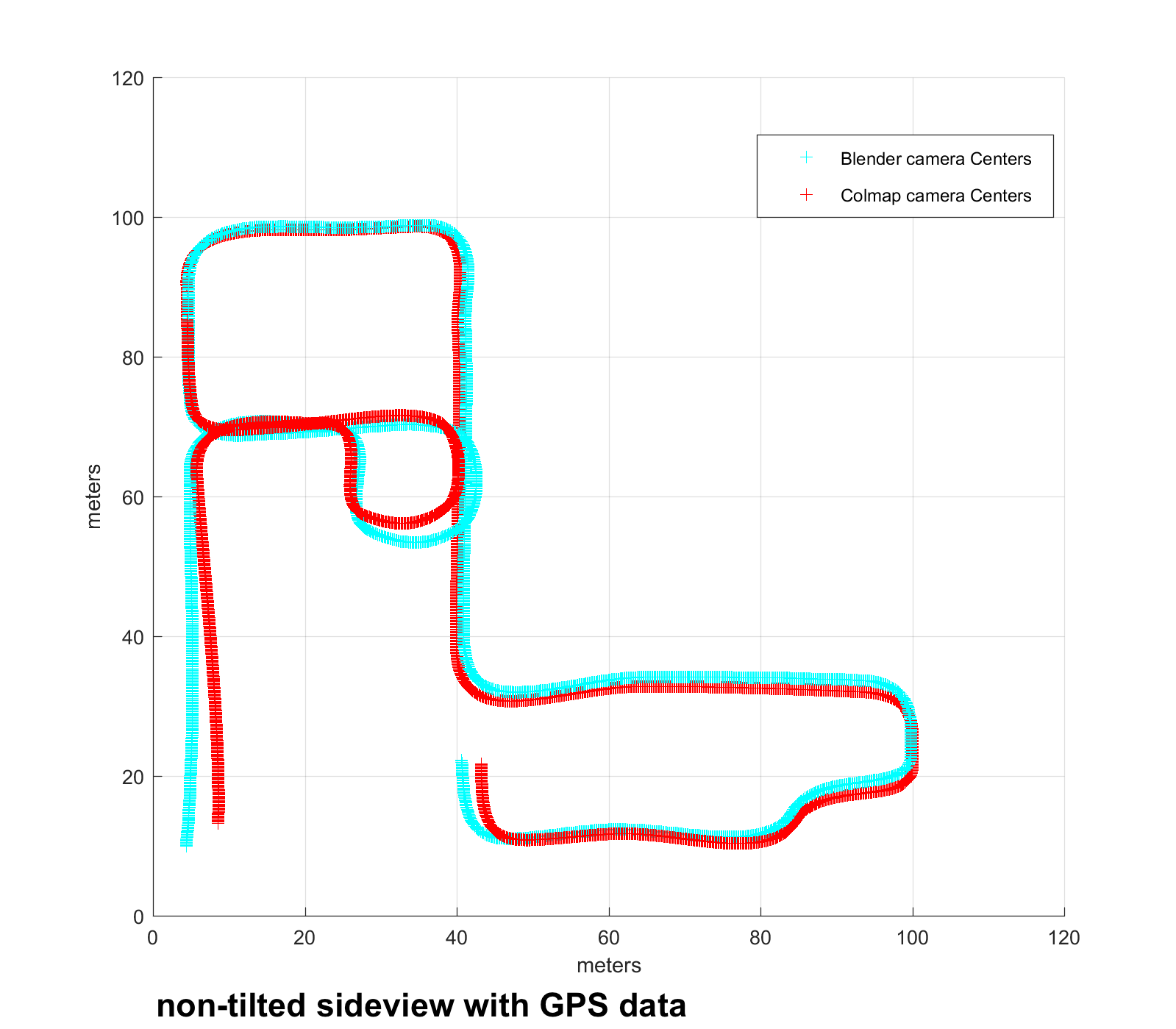}
   		\includegraphics[width=0.49\linewidth]{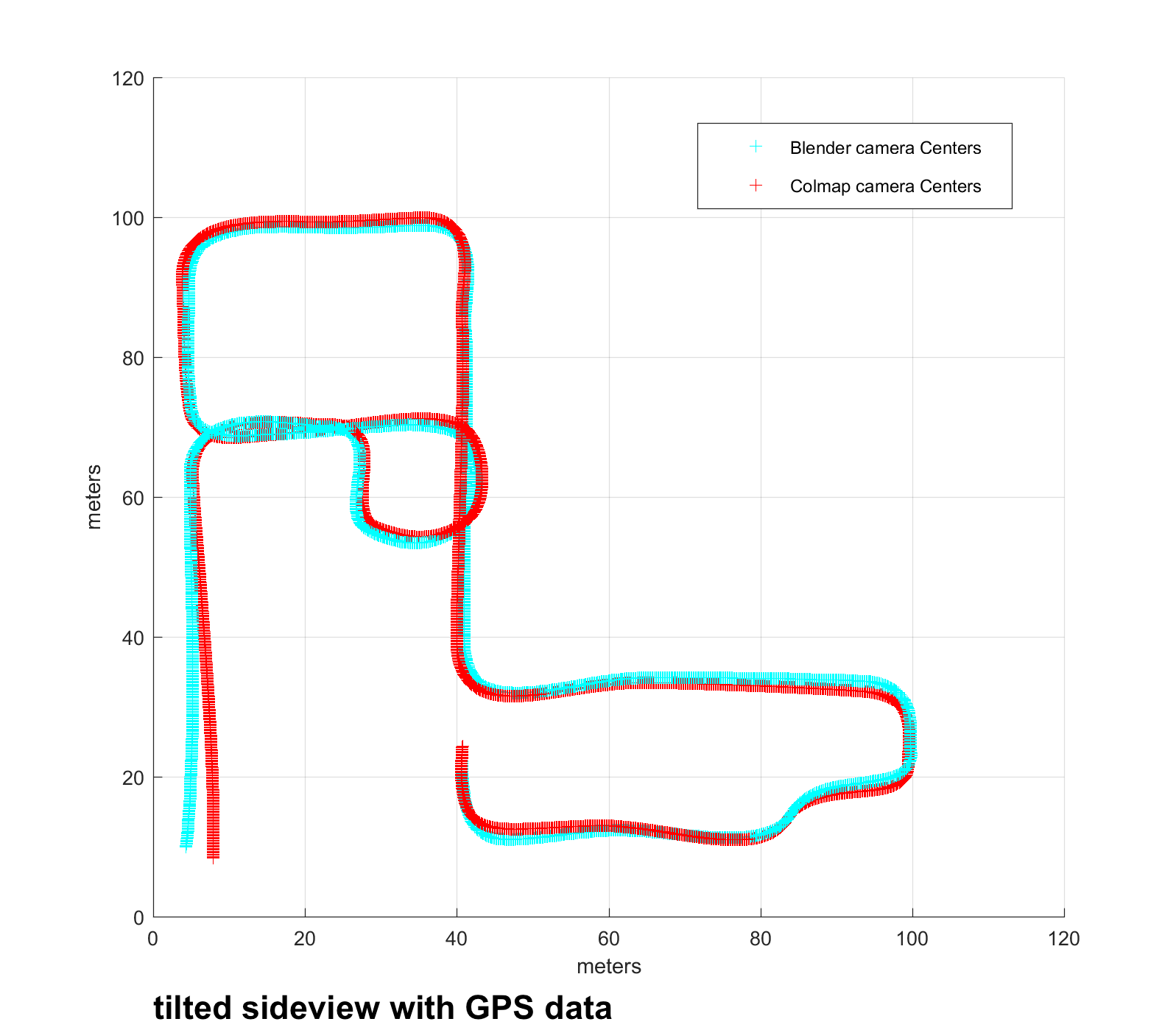}
   		\caption{camera poses of (c): non-tilted sideview with GPS data, (d) tilted sideview  with GPS data}
	\end{center}
\end{figure}

Lastly, SfM reconstructions with spatial feature matching, only two neighboring images and presumable better feature criteria as described in section~\ref{SFMspat} were computed and a comparison of the camera poses from blender and COLMAP are presented in figure 11. 

\begin{figure}[ht] \label{cameraposesef}
	\begin{center}
      	\includegraphics[width=0.49\linewidth]{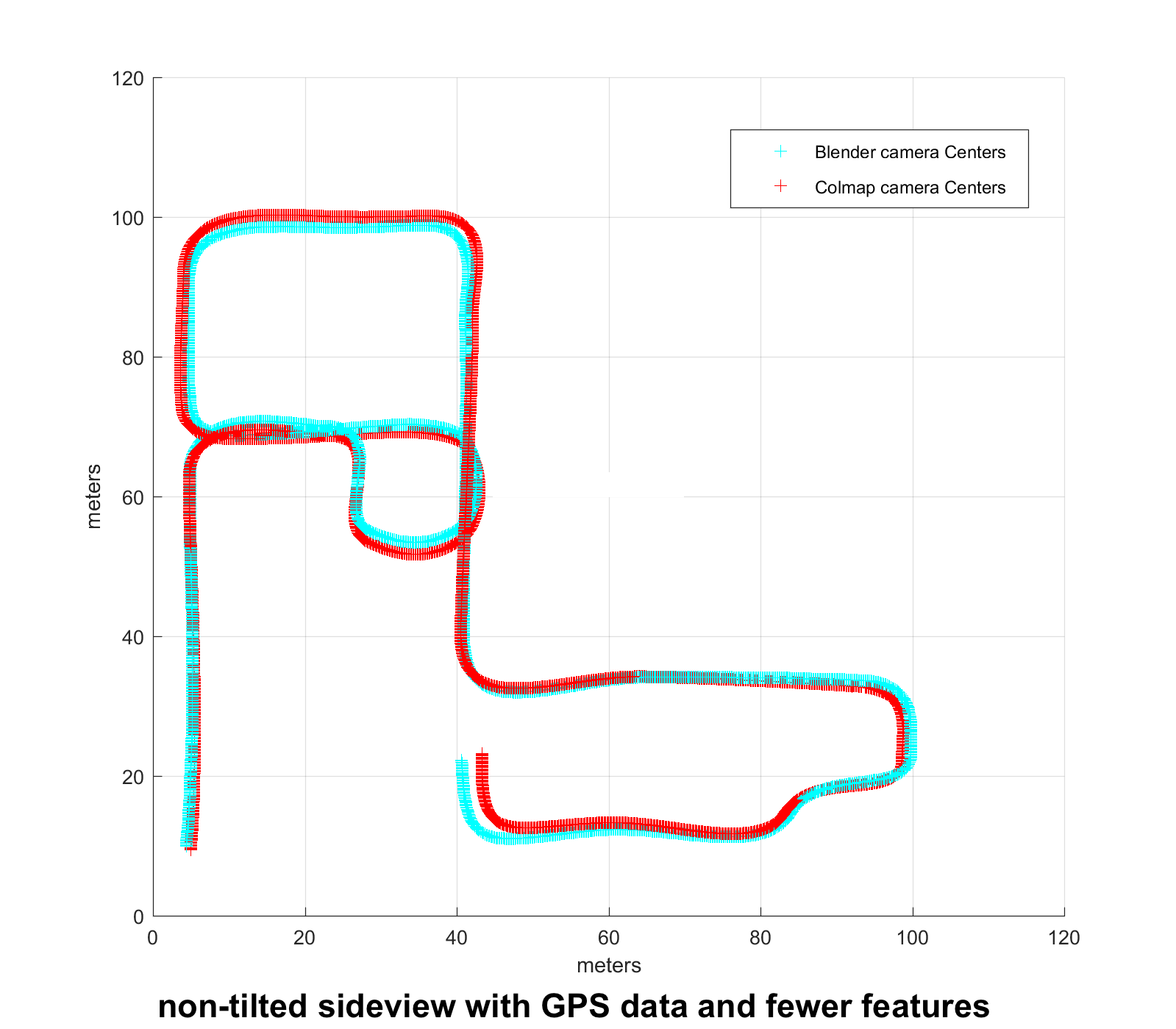}
   		\includegraphics[width=0.49\linewidth]{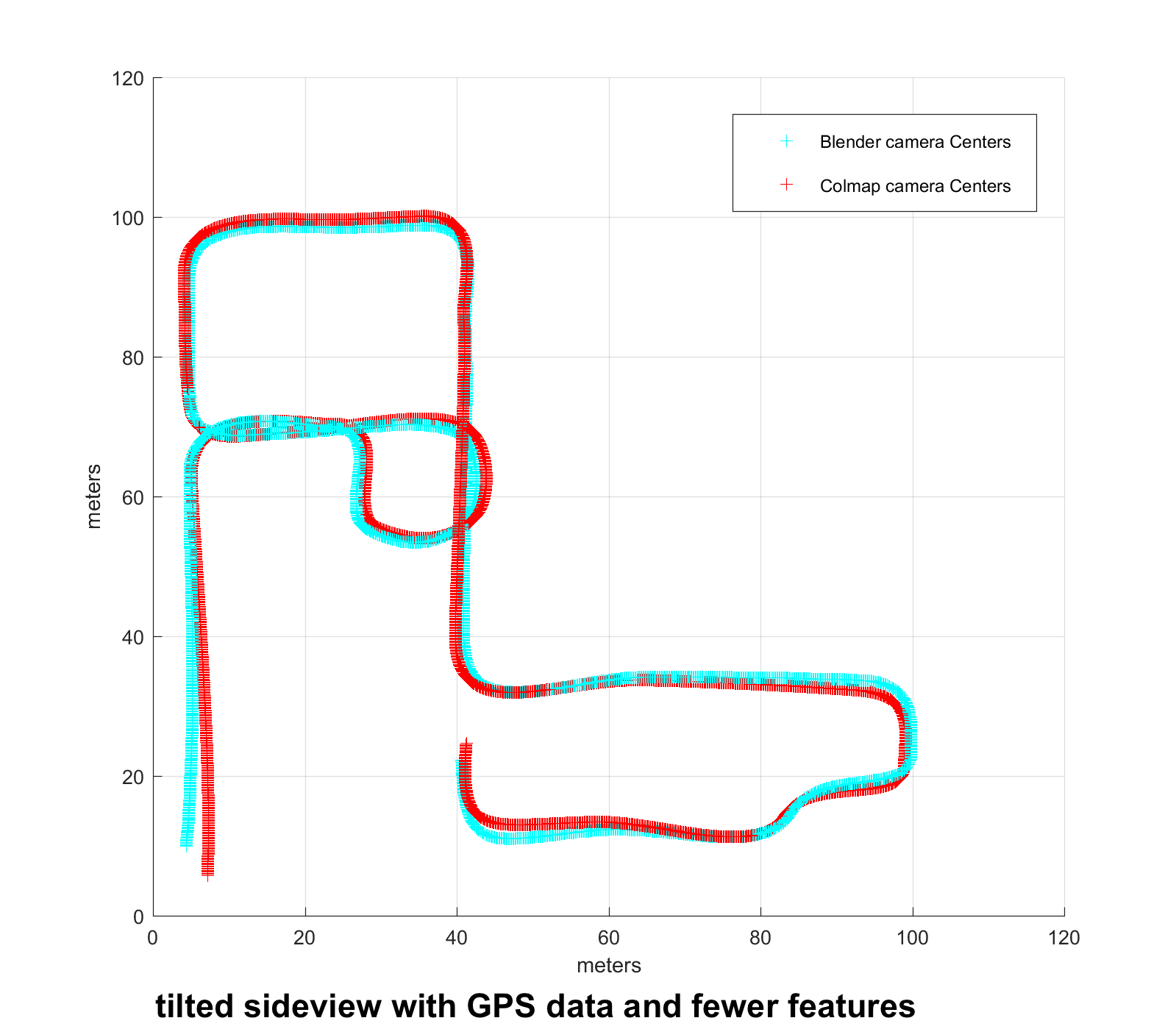}
   		\caption{camera poses of (e): non-tilted sideview with GPS data and fewer features, (f): tilted sideview with GPS data and fewer features}
	\end{center}
\end{figure}

All the plots of this section were generated with Matlab after following the steps introduced in section~\ref{Calib}. However, since it is hard to qualitatively compare those different reconstructions visually, the reconstructions also were validated using Matlab with the error metric explained in the following section.

\subsection{Validation of the different SfM reconstructions} \label{Results-Validation}
At this point, the validation of the SfM reconstructions is necessary in order to assess the practical use of the proposed pipeline. The plots of the reconstructions shown above provide a great insight of the overall faithfulness of the representations, but now a more thorough examination of the results using statistical tools is presented.

The root-mean-square (RMS) error in all three directions $x$, $y$ and $z$ as well as an average RMS error between the camera poses extracted from COLMAP and the ground truth camera poses from blender were investigated. The RMS of every SfM reconstruction mentioned in this paper are shown in figure 12. 

\begin{figure}[ht] \label{RMS}
	\begin{center}
       	\includegraphics[width=1\linewidth]{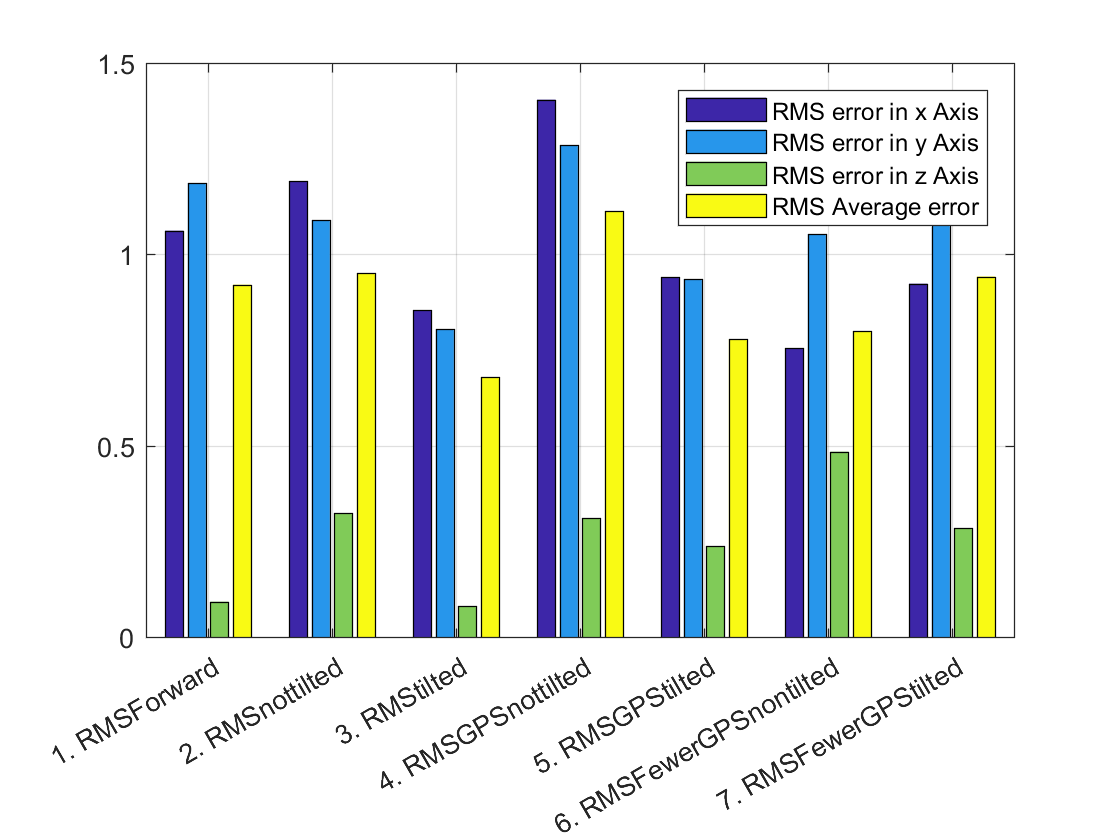}
   		\caption{RMS for the $x$, $y$ and $z$ direction as well as the average RMS for the camera pose calibration}
	\end{center}
\end{figure}

For completeness, the equation that was used and implemented in Matlab to compute the RMS is stated:

\begin{equation*}\label{RMSerror}
	RMS = \sqrt {{\frac{{\sum\limits_{{i = 1}}^n {{{\left( {{y_i} - {{\hat{y}}_i}} \right)}^2}} }}{{n}}}}
\end{equation*}

For all cases in the sequential feature matching reconstructions, the RMS average error of the camera poses was less than $1m$, which proves the accuracy of the reconstructions bearing in mind the dimensions of the whole synthetic city scene were around $100m$ x $100m$. The individual error in the $z$ direction was almost negligible since in most cases it was less than $0.3m$. The biggest errors of the poses were observed in the planar directions $x$ and $y$ which were around $1m$. The best reconstructions were computed using the tilted sideview camera orientation.

The results in case of spatial feature matching reconstructions with ten neighboring images were slightly worse as the errors in all directions were increased by approximately $0.15m$, except for the $z$ direction which seemed to be more robust and remained almost the same. An interesting phenomenon was observed when only two neighboring images and presumable better feature criteria as described in section~\ref{SFMspat} were used. In this case the results for the tilted sideview camera orientation were deteriorated slightly and an improvement only took place for the non-tilted sideview orientation of the camera. In this case, the significant improvement was made in the $x$ direction, since the error plunged to almost half of the error that was observed before. Again, in spatial feature matching reconstructions the best results were obtained for the tilted sideview camera orientation. The sorted average RMS error overview is shown in figure 13.

\begin{figure}[ht] \label{RMSAVG}
	\begin{center}
       	\includegraphics[width=1\linewidth]{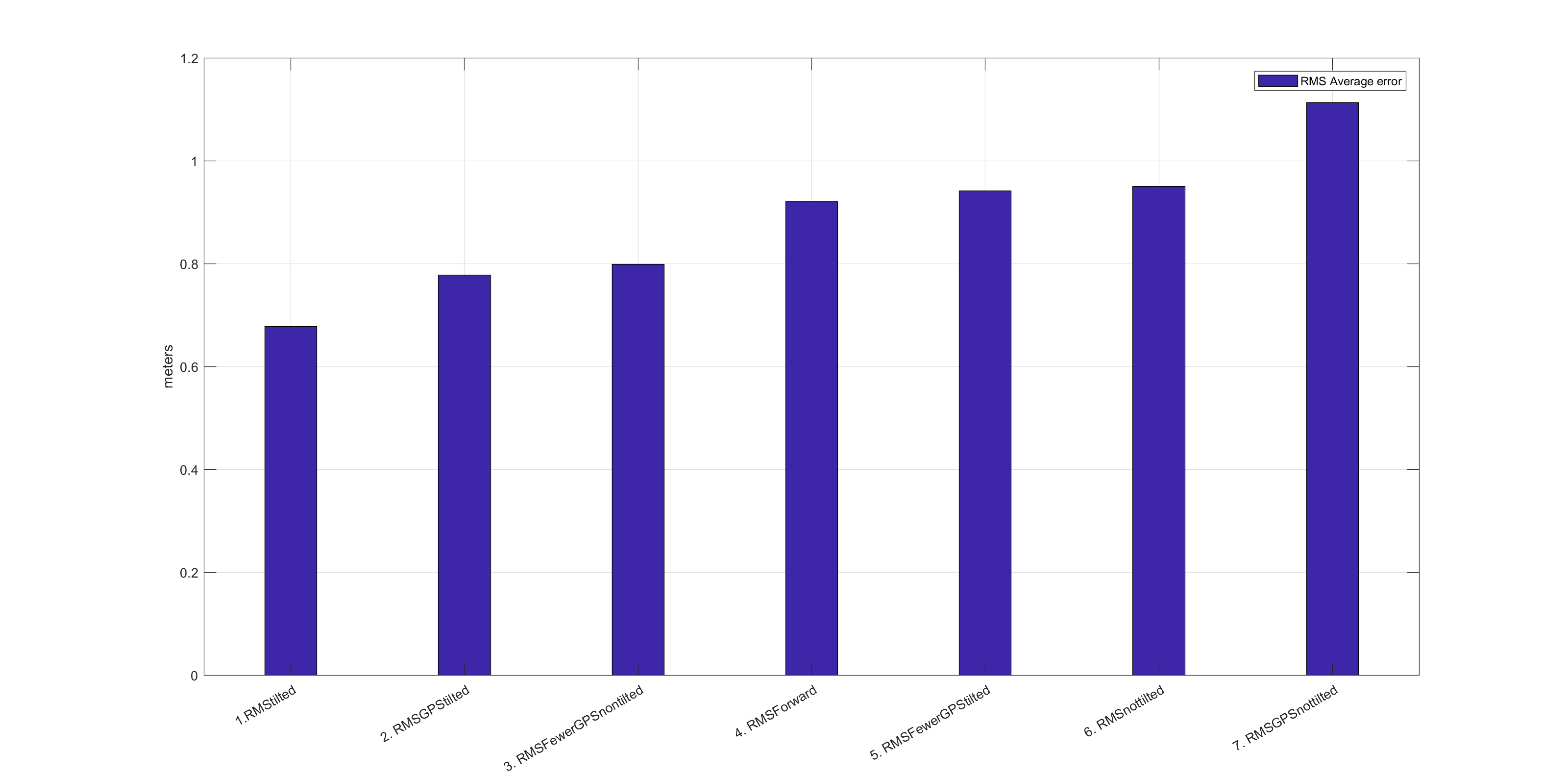}
   		\caption{sorted average RMS for the camera pose calibration}
	\end{center}
\end{figure}

An explanation for the unexpected overall worse camera poses for the case of the spatial feature matching reconstructions after inserting  the ground truth GPS data from blender into the EXIF location data of each image in COLMAP is the following. COLMAP uses this inserted ground truth information as a prior for the optimization problem that solves to extract the posterior camera poses. As a result, it seems that no matter if an initial guess is inserted to the solver of COLMAP, the program will eventually end up to different results due to several global bundle adjustment iterations.

\subsection{Dense Reconstruction} \label{Results-Dense}
Finally, for visualization purposes and after investigation of the error metric introduced in the previous section, the sparse SfM reconstruction with tilted images was chosen for the conduction of a dense reconstruction. Sparse reconstruction, basically is a 3D point cloud from the common matches between a set of images. COLMAP and other SfM reconstruction packages such as VisualSFM produce the sparse reconstruction and put a color in each 3D point similar to the one found on the 2D points in the images it corresponds. The sparse SfM reconstruction is a segmentation of the volume of a real object into 3D points with still a lot of empty space between those 3D points. The dense reconstruction, which is also called Multi-View-Stereo, comes to bridge the gap of the empty space between the 3D points of the point cloud. This is done by integrating the resulting point cloud into an energy functional by taking the output of the spares SfM reconstruction and finding depth information for every pixel in the rendered images. Fusing the depth maps of all the images in 3D by using algorithms such as the poison surface reconstruction, 3D geometry can be recovered~\cite{schoenberger2016mvs}. Four snapshots of the resulting dense reconstruction are shown in figure 14. It should be mentioned that for this dense reconstruction a cluster computer from the CVG lab was used due to the highly burdensome procedure. This procedure always crashed on the workstation used to render the images and compute the sparse SfM reconstructions due to not enough RAM. An animation of the dense reconstruction can be found on
\MYhref{https://vimeo.com/martinhahner/denseanimation}{https://vimeo.com/martinhahner/denseanimation}.  

\begin{figure}[ht] \label{dense}
	\begin{center}
       	\includegraphics[width=0.45\linewidth]{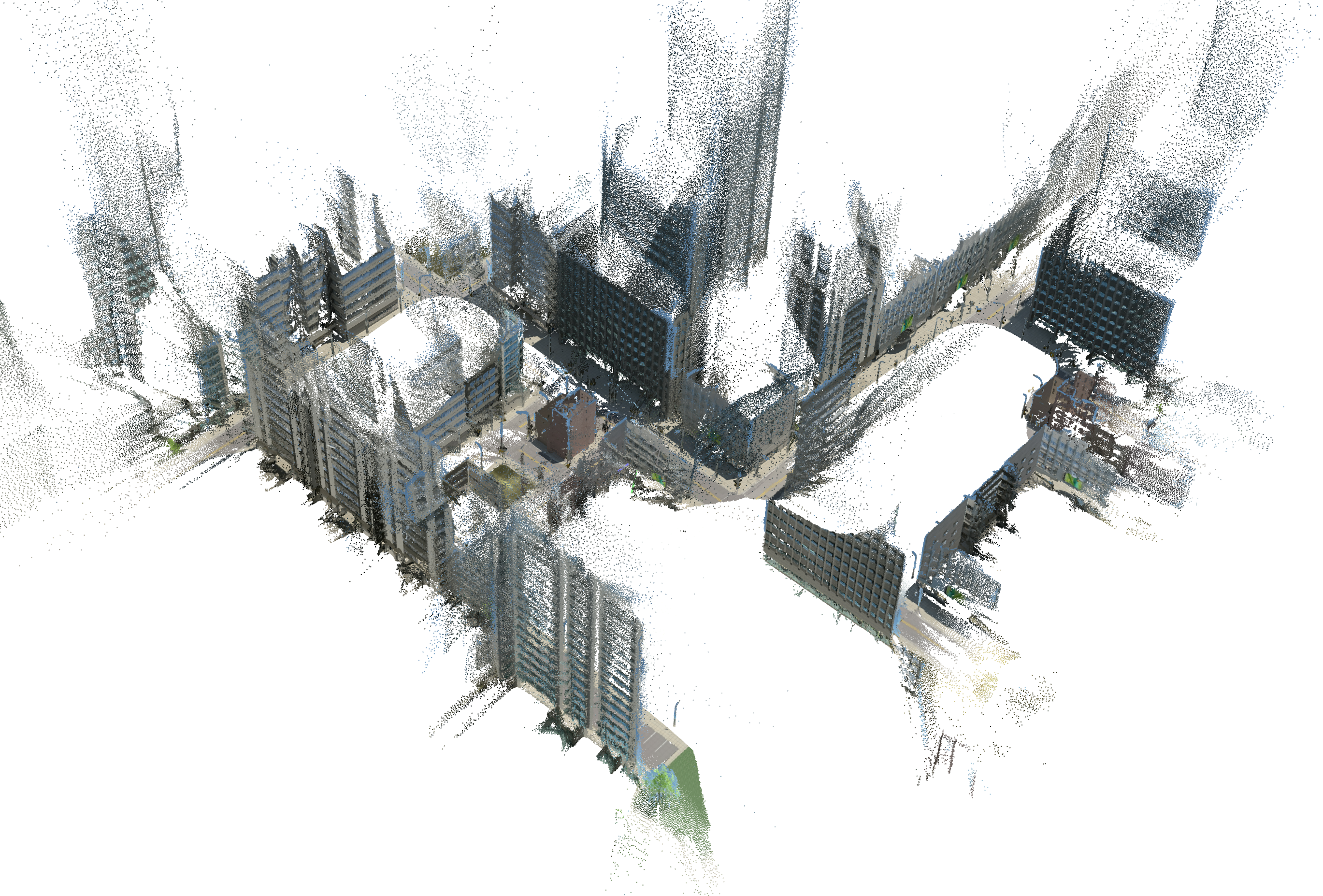}
       	\includegraphics[width=0.45\linewidth]{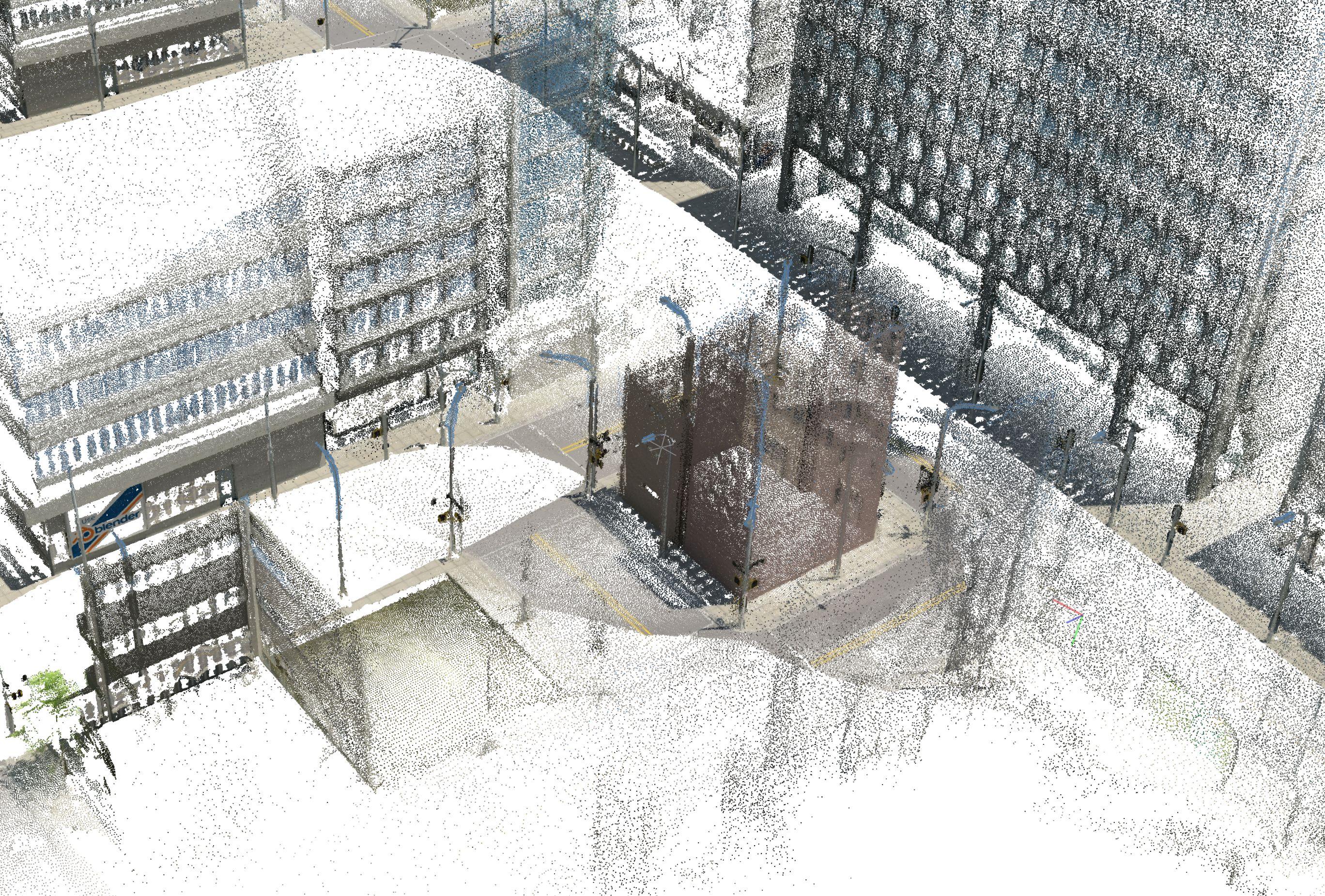}
        \includegraphics[width=0.45\linewidth]{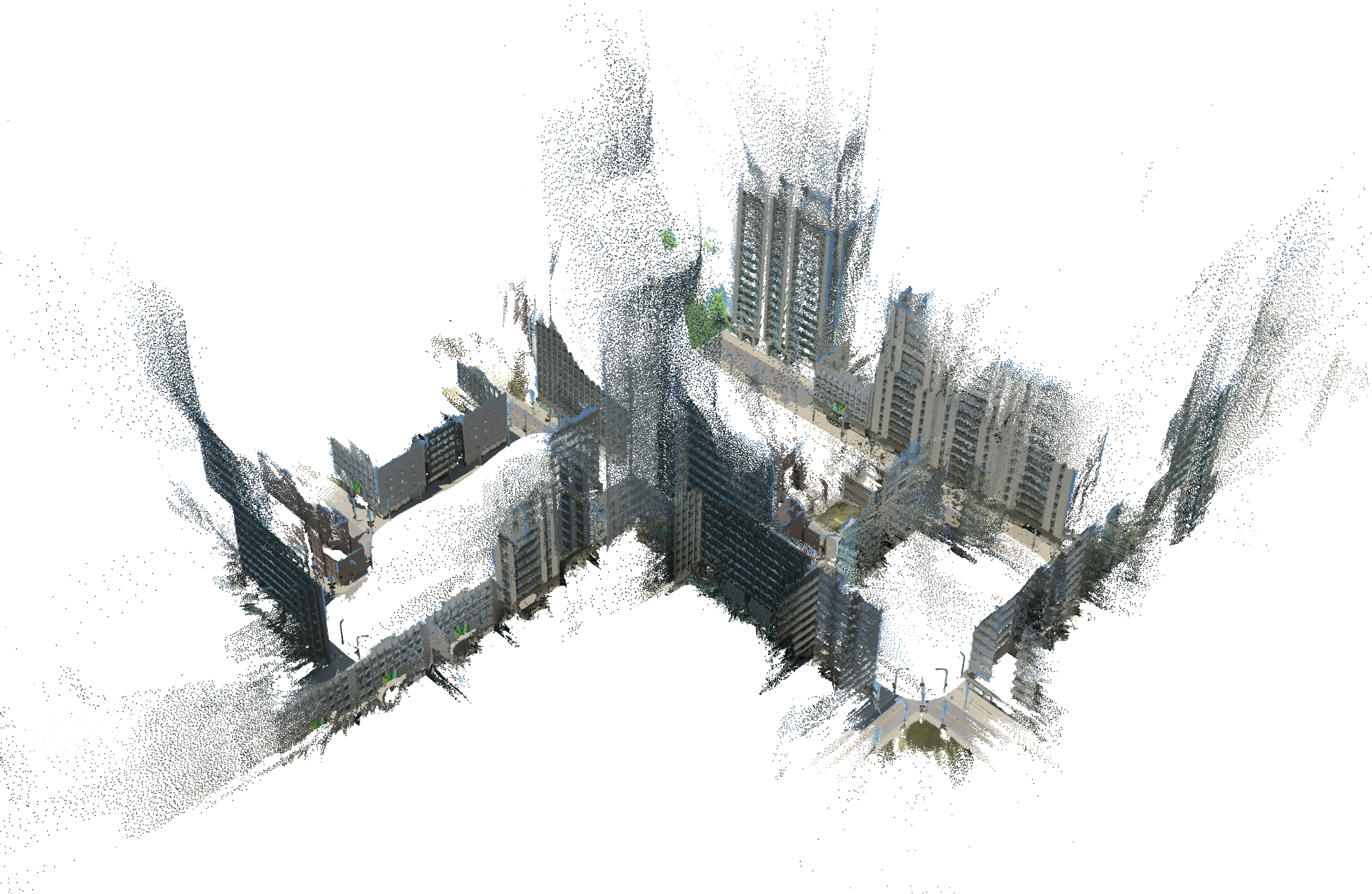}
       	\includegraphics[width=0.45\linewidth]{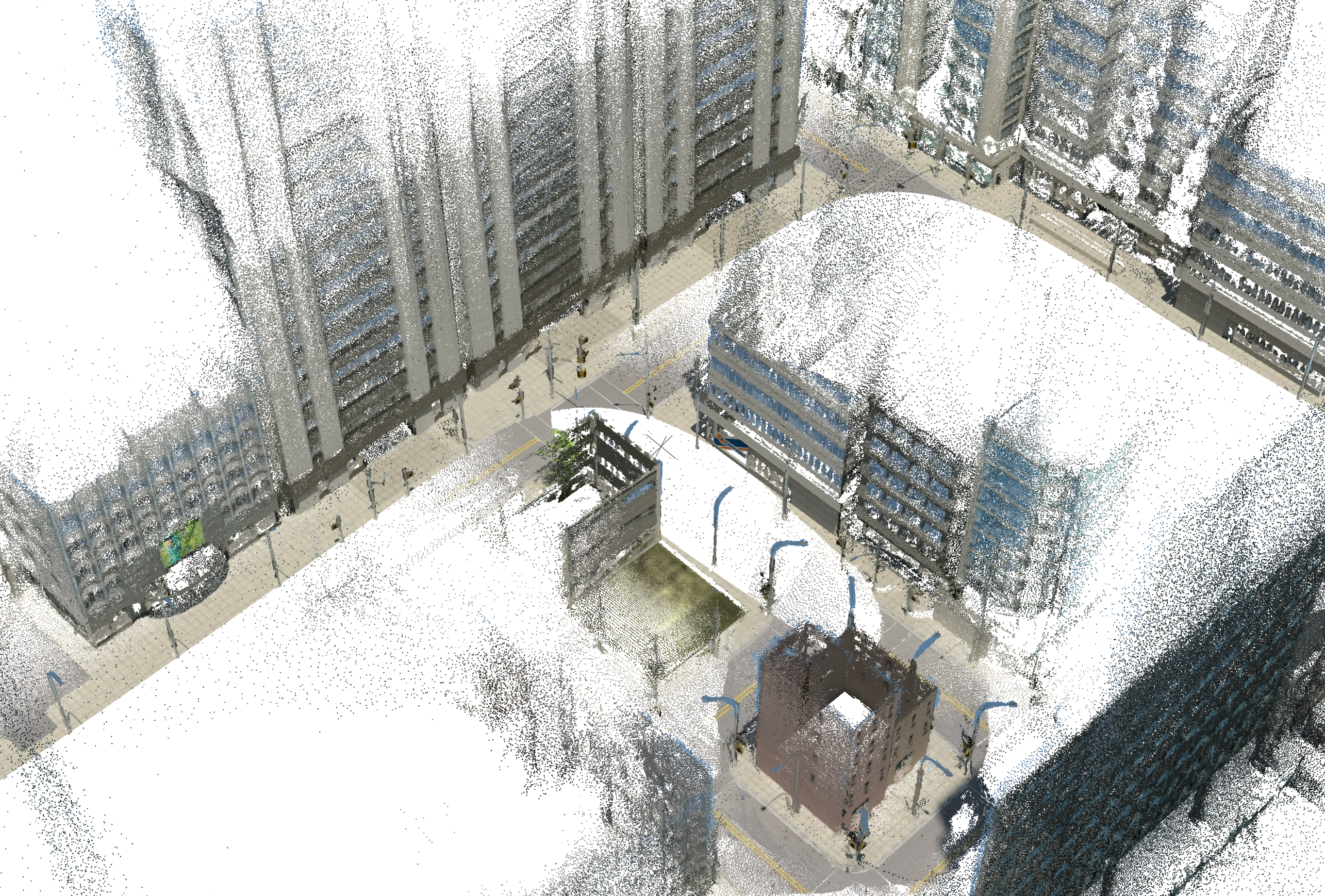}
   		\caption{snapshots of our best dense SfM reconstruction using the tilted sideview images and sequential feature matching with 10 sequential images}
	\end{center}
\end{figure}

\section{Conclusions} \label{Conclusions}
Summarizing, after analyzing the key aspects of this project, the camera orientation which is proposed to be used in rendering images is a tilted sideview. In regards to COLMAP, the sequential feature matching seems to result in more faithful representations and camera poses than the spatial feature matching. 

With the proposed pipeline it is possible to compare different SfM reconstructions from synthetically generated data with precise ground truth and determine the accuracy of the underlying algorithms. It is also possible to investigate other properties of those algorithms, for example  how noise effects them or how many images are necessary for a reasonable good reconstruction.

One important hint that was drawn from our involvement in this project and we would like to share this with those interested in the field, is which steps should be taken in case the SfM reconstruction fails. If someone faces this problem in the future, then the first action one should do is to reduce the actual space between each rendered image so that more information is inserted to the algorithms and better feature matching is achieved. Indicatively, in our project when only every tenth frame along the camera trajectory was used the SfM reconstruction failed, especially in the corners of blocks, since not adequate feature matches were available and therefore renderings of every single frame were used. 

In the extreme case that this action would not solve the problem, an additional step that could be made would be to increase the texture of the synthetic scene. The goal of this step is to further increase the information of the scene provided to the algorithms and amend the feature matching. However, it should be stated that this would dramatically increase the computation time for both the image renderings and the SfM reconstructions. As a result, a trade-off between the desired texture and the computation time which is needed is inevitable.

\section{Future Work} \label{FutureWork}
In a subsequent project it could be investigated how faithful the 3D point reconstruction would be if COLMAP is forced to keep the initial ground truth camera locations and is not allowed to diverge from them. In a first step one could check if a single global bundle adjustment with injected ground truth, which is run after COLMAP has finished the SfM reconstruction, could lead to slightly better results.

{\small
\bibliographystyle{ieee}
\bibliography{egbib}
}

\end{document}